\documentclass[letter, 10 pt, conference]{ieeeconf}
\IEEEoverridecommandlockouts                              
\usepackage{times}
\usepackage{epsfig}
\usepackage{graphicx}
\usepackage{amsmath}
\usepackage{amssymb}
\usepackage{multirow}

\usepackage{graphicx}
\usepackage{epsfig}
\usepackage{epic,eepic}
\usepackage{times}
\usepackage{mathptmx}
\usepackage{amsfonts}
\usepackage{amsmath}
\usepackage{cite}
\usepackage{color}

\usepackage{times}

\definecolor{red}{rgb}{1,0,0}
\definecolor{green}{rgb}{0,1,0}
\definecolor{blue}{rgb}{0,0,1}
\definecolor{violet}{rgb}{1,0,1}
\definecolor{cyan}{cmyk}{1,0,0,0}
\definecolor{magenta}{cmyk}{0,1,0,0}
\definecolor{yellow}{cmyk}{0,0,1,0}

\definecolor{white}{rgb}{1,1,1}

\usepackage{jumoline}

\newcommand{\CO}[2]{}

\newcommand{\CommentOut}[1]{}

\newcommand{\FIG}[3]{
\begin{minipage}[b]{#1cm}
\begin{center}
\includegraphics[width=#1cm]{#2}\\
{\scriptsize #3}
\end{center}
\end{minipage}
}

\newcommand{\FIGS}[3]{
\begin{minipage}[b]{#1cm}
\begin{center}
\includegraphics[bb=0 0 600 787, clip, height=4cm, width=4cm]{#2}\\
{\scriptsize #3}
\end{center}
\end{minipage}
}

 \newcommand{\editage}[1]{}
\onecolumn

\renewcommand{\CO}[2]{#2}

\newcommand{\tabD}{
\begin{table}[t]
\begin{center}
\caption{Performance results in Top-$X$ accuracy [\%].}\label{tab:D}
\begin{tabular}{|r|r|r|r|r|r|r|}
\hline
\multirow{2}{3mm}{$X$} & \multicolumn{2}{c|}{rAE} & \multicolumn{2}{c|}{k-means} & \multicolumn{2}{c|}{NN} \\ \cline{2-7}
& $\ge$25 & $\ge$50 & $\ge$25 & $\ge$50 & $\ge$25 & $\ge$50 \\ \hline
5 & 7.5 & 2.4 & 7.3 & 1.9 & 0.0 & 0 \\
10 & 17.8 & 5.5 & 17.1 & 3.7 & 3.2 & 0\\
15 & 29.3 & 10.5 & 28.6 & 8.3 & 22.0 & 0 \\
20 & 41.3 & 14.7 & 40.3 & 13.2 & 46.0 & 1.0\\
\hline
\end{tabular}
\end{center}
\end{table}
}

\newcommand{\figA}{
\begin{figure}[t]
    \begin{minipage}[b]{16.5cm}
  \begin{center}
  \FIG{16}{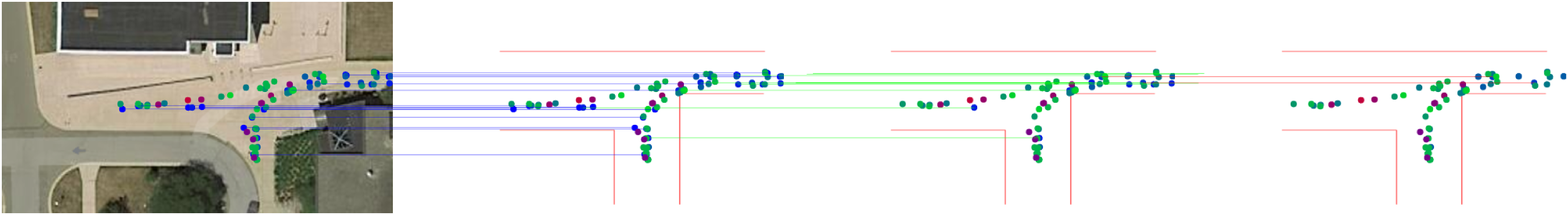}{}
\vspace*{-5mm}
 \end{center}
    \caption{
Recursive background modeling.
Each panel shows
from left to right,
viewpoints of 
all background images,
viewpoints of 
background images
that are not well explained by 
the first,
second,
and third autoencoder (AE),
respectively.
Colored horizontal line segments
connect
the viewpoints 
explained by each AE.
}\label{fig:A}
    \end{minipage}
\end{figure}
}

\newcommand{\figC}{
\begin{figure}[t]
\begin{center}
\begin{minipage}[b]{4cm}
\FIG{4}{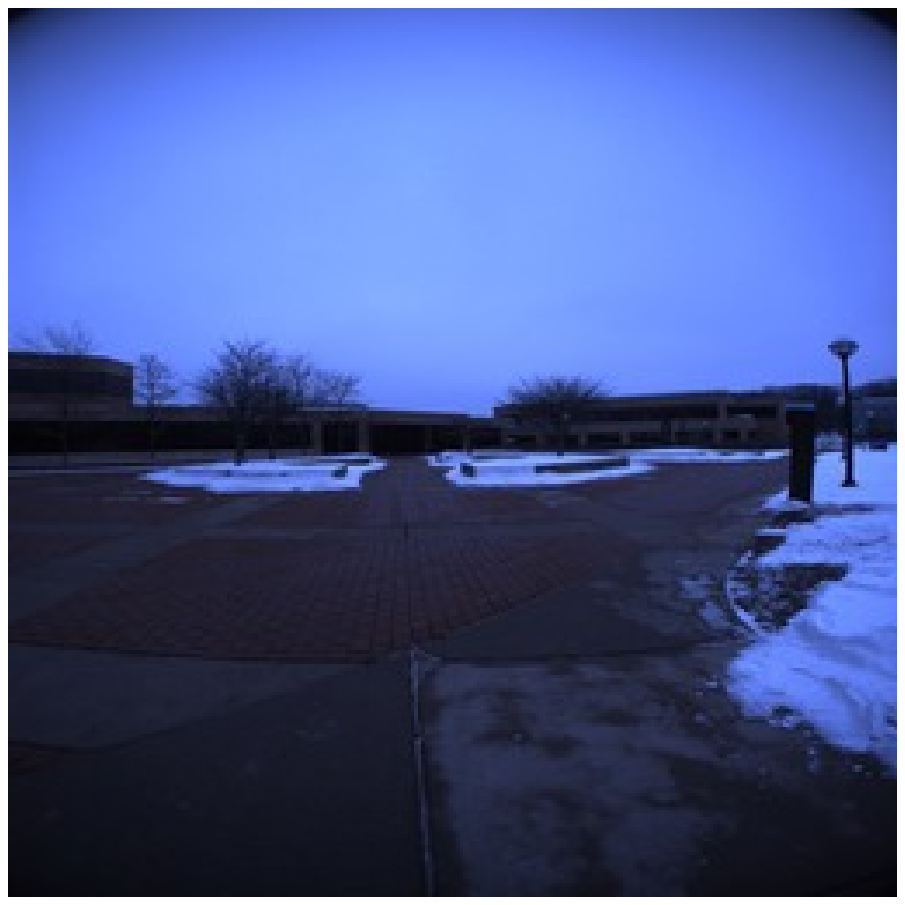}{}\vspace*{2mm}\\
\hspace*{-3mm}\FIGS{4}{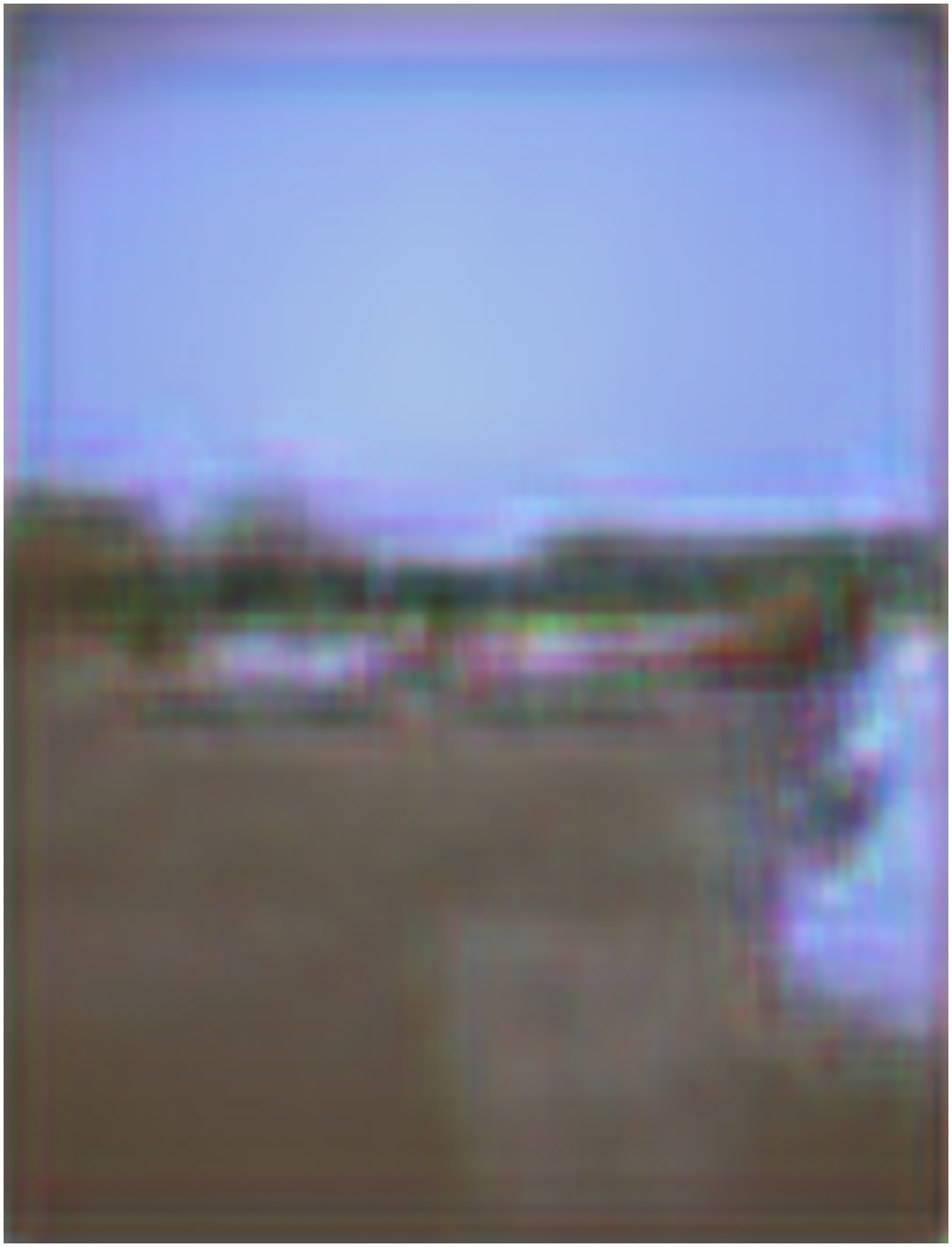}{} \\
{\scriptsize (a) RE = 1.2907} 
\end{minipage}\hspace*{10mm}%
\begin{minipage}[b]{4cm}
\FIG{4}{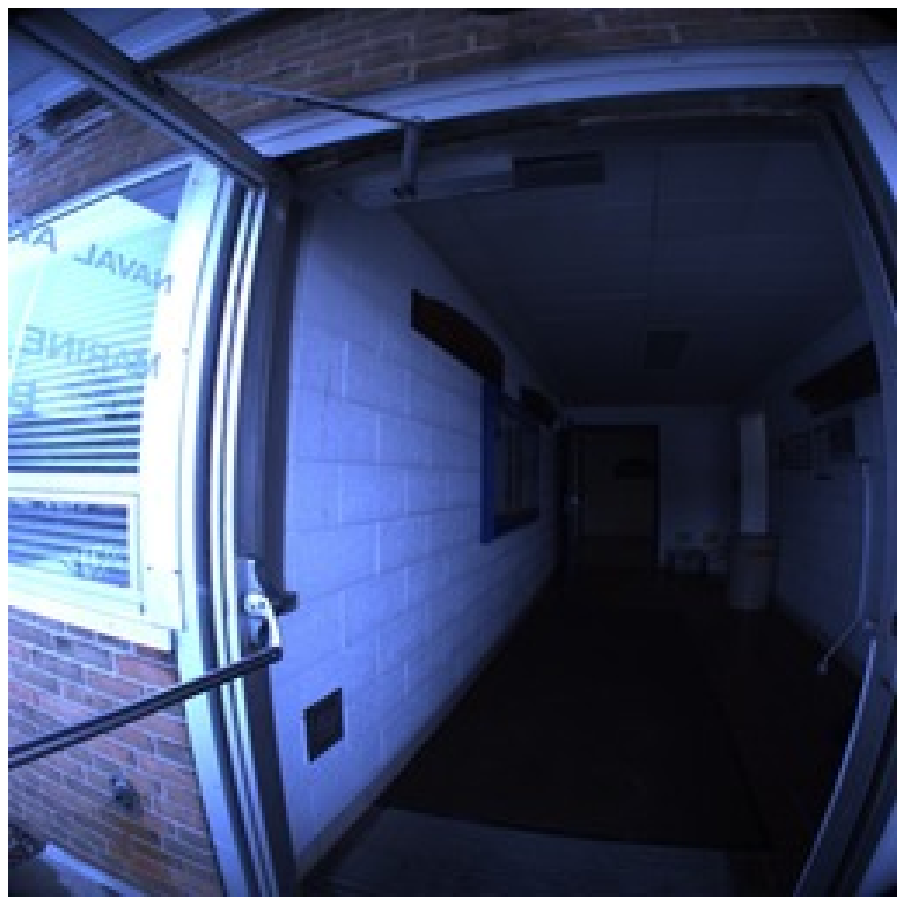}{}\vspace*{2mm}\\
\hspace*{-3mm}\FIGS{4}{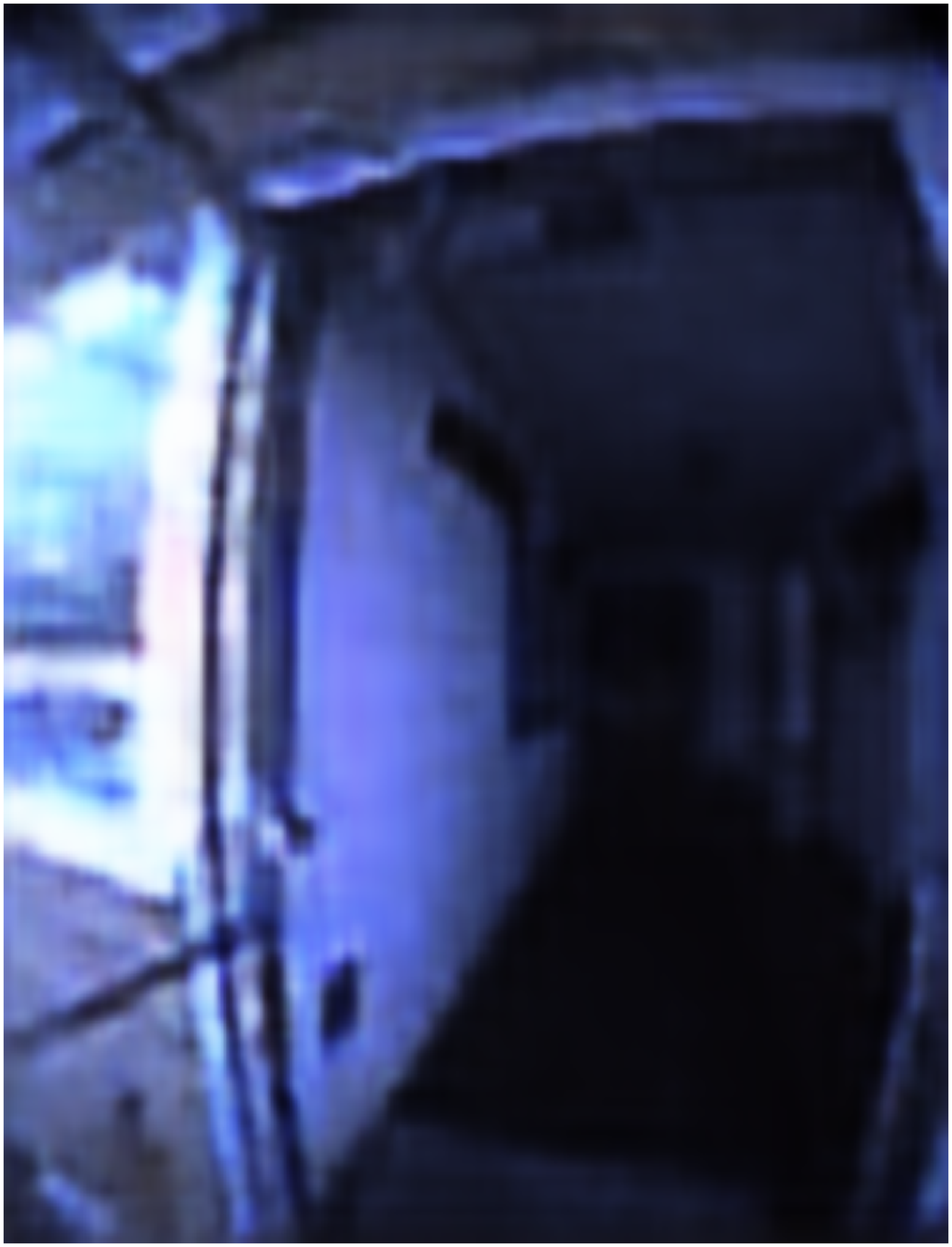}{} \\
{\scriptsize (b) RE = -0.0335}
\end{minipage}\hspace*{10mm}%
\begin{minipage}[b]{4cm}
\FIG{4}{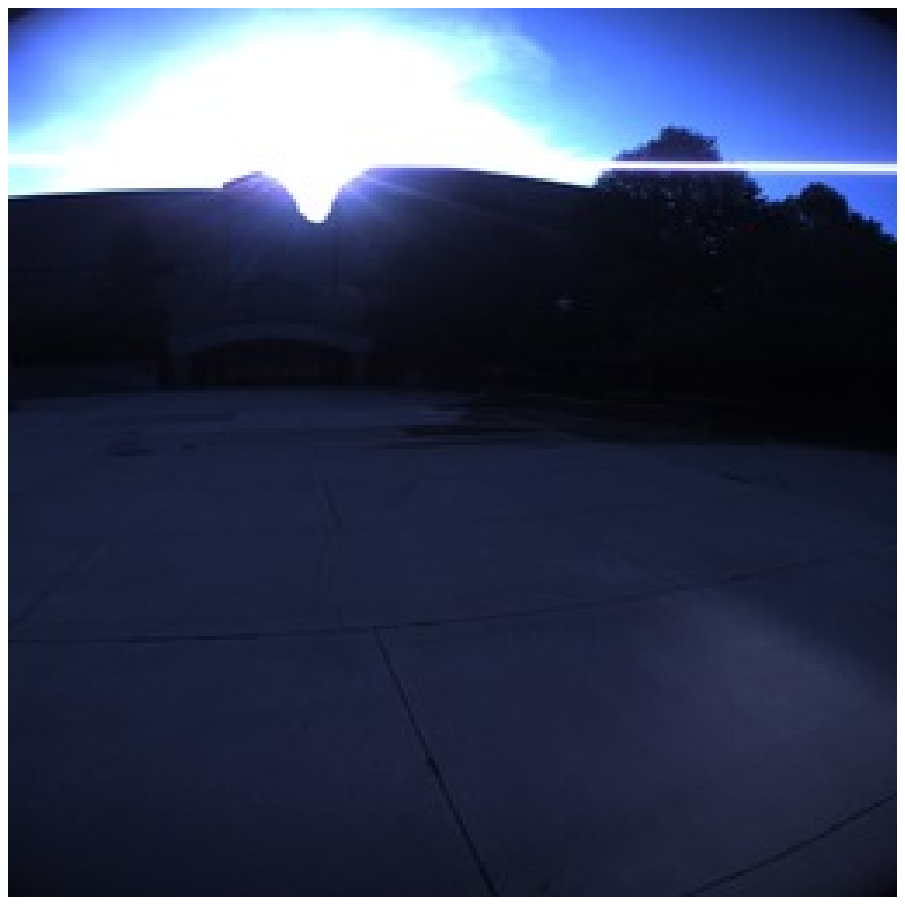}{}\vspace*{2mm}\\
\hspace*{-3mm}\FIGS{4}{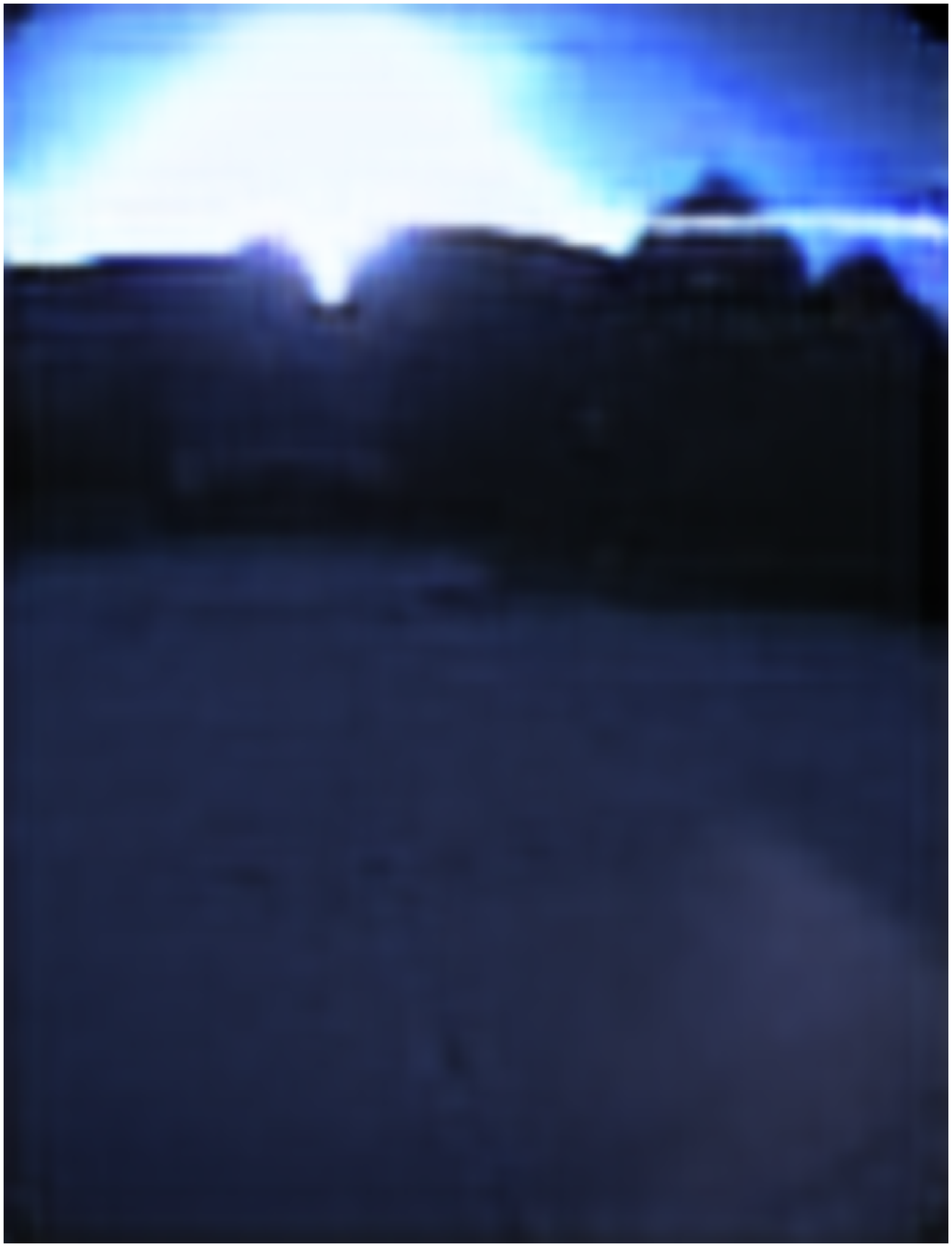}{} \\
{\scriptsize  (c) RE = -0.5431} 
\end{minipage}
\caption{Examples of input and reconstructed image pairs 
and the normalized reconstruction errors.}\label{fig:C}
 \end{center}
\end{figure}
}

\newcommand{\figD}{
\begin{figure}[t]
 \begin{center}
  \FIG{5}{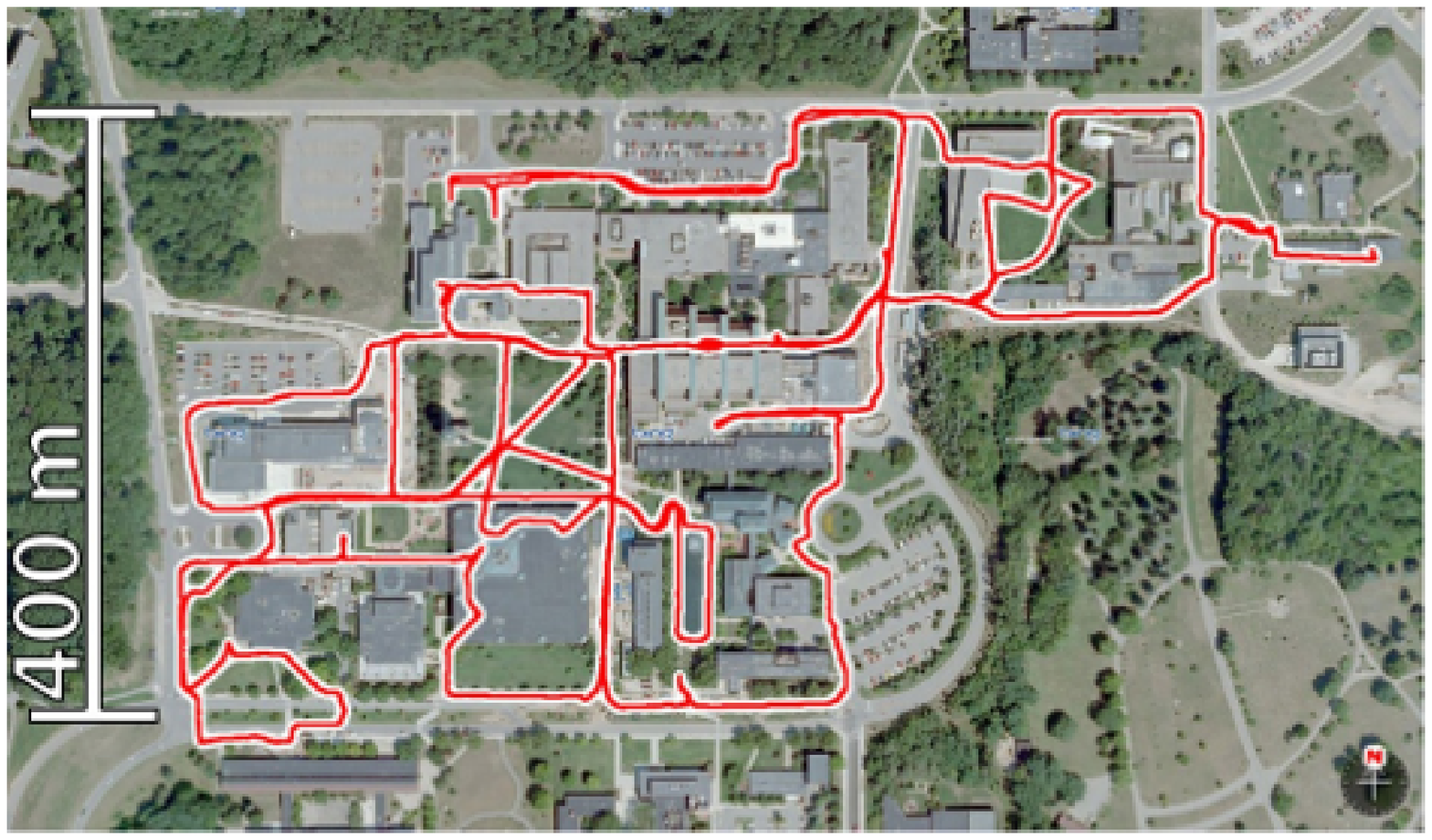}{}
 \end{center}
 \caption{Experimental environment and vehicle trajectories.} \label{fig:nclt}
\end{figure}
}

\begin{document}

\title{\LARGE \bf
Place-specific 
Background Modeling 
Using
Recursive Autoencoders
}

\author{Yamaguchi Kousuke ~~~~ Tanaka Kanji ~~~~ Sugimoto Takuma ~~~~ Ide Rino  ~~~~ Takeda Koji
\thanks{Our work has been supported in part by JSPS KAKENHI Grant-in-Aid for Scientific Research (C) 26330297, and (C) 17K00361.}
\thanks{The authors are with Graduate School of Engineering, University of Fukui, Japan.
{\tt\small tnkknj@u-fukui.ac.jp}}
}

\maketitle

\section*{\centering Abstract}
\textit{
Image change detection (ICD) to detect changed objects in front of a vehicle with respect to a place-specific background model using an on-board monocular vision system is a fundamental problem in intelligent vehicle (IV). From the perspective of recent large-scale IV applications, it can be impractical in terms of space/time efficiency to train place-specific background models for every possible place. To address these issues, we introduce a new autoencoder (AE) based efficient ICD framework that combines the advantages of AE-based anomaly detection (AD) and AE-based image compression (IC). We propose a method that uses AE reconstruction errors as a single unified measure for training a minimal set of place-specific AEs and maintains detection accuracy. We introduce an efficient incremental recursive AE (rAE) training framework that recursively summarizes a large collection of background images into the AE set. The results of experiments on challenging cross-season ICD tasks validate the efficacy of the proposed approach.
}

\section{Introduction}

Image change detection (ICD) for detecting changed objects in front of a vehicle as compared with a place-specific background model using an on-board monocular vision system is a fundamental problem in intelligent vehicle (IV). Many researchers recently addressed IV applications in large-scale real-world scenarios (e.g., intelligent transportation systems) rather than small laboratory scenarios \cite{nclt}. For these scenarios, to train place-specific background models for every possible place can be impractical. It requires very considerable space/time costs proportional to the environment size. The motivation of this study was to enhance model compactness while maintaining the effectiveness of the ICD system.

As a primary contribution, we introduce a new autoencoder (AE) based efficient ICD framework that combines the advantages of AE-based anomaly detection (AD) and AE-based image compression (IC). An AE is an unsupervised neural network that takes an input image and maps it to a hidden representation using an encoder network and then uses a decoder network to reconstruct the input from the hidden representation. Our AE-based approach is motivated by recent successes in two independent research fields: AD \cite{ccr2} and IC \cite{balle2016end}. (1) In the field of learning-based AD, it has recently been reported that the reconstruction error of an AE is a very good indicator of anomalous or changed objects \cite{HinSal06}. (2) In the field of learning-based IC, the employment of an AE as a compressor/decompressor of input/output images has become a de facto standard \cite{torfason2018towards}.

We were particularly interested in the use of place-specific AEs as a basis of ICD and explored their use in this study. Suppose a given sequence of place-specific background images acquired by a vehicle along a known trajectory in the target environment. We aim to partition the background image collection into $N$ disjoint clusters and train a minimal collection of $N$ place-specific AEs $\{c_1$, $\cdots$, $c_N\}$ from individual clusters. A naive solution would be to partition the image sequence into equal-length subsequences and then train each AE using each subsequence. However, this solution ignores the appearance of background images and their semantic similarity, which may lead to a suboptimal detection/compression performance.

\figA

To address the above issues,
we propose
a novel recursive AE (rAE)
training method
that 
exploits
the reconstruction error (RE) of AEs 
as a key measure (Fig. \ref{fig:A}).
That is,
we perform 
{\it virtual AD}
using the background image as input to the current AE set.
This virtual AD provides two important cues (Fig. \ref{fig:C}).
(1)
If the RE is larger than the allowable error,
it is likely that 
the background image cannot be modeled by the current AE set,
indicating the necessity of adding a new AE (i.e., $N_{i+1}$$\leftarrow$$N_i+1$).
(2)
If the RE is sufficiently small for an existing member $c_j$$(j\in [1,N_i])$ of the current AE set,
it can then be used
as training data to update
the binary normal/anomalous decision boundary of that AE $c_j$
(i.e., $N_{i+1}$$\leftarrow$$N_i$).
Thus,
our approach exploits
the RE
as a single unified measure for
minimizing the number of AEs
while maintaining the detection accuracy of individual AEs.
After the $i$-th AE $c_i$ has been trained from a training image set $T_i^+$,
the training set is classified by the current AE set $\{c_j\}_{j=1}^i$ into
a normal subset $T_{i+1}^-$ and an anomalous subset $T_{i+1}^+$,
and then the latter set is used to train the next generation AE $c_{i+1}$ recursively.
We implemented the proposed algorithm and verified its effectiveness in a challenging scenario of cross-season ICD using 
the publicly available 
North Campus Long-Term
(NCLT) dataset \cite{nclt}.

\figC

\section{Approach}

\subsection{Problem Formulation}

We consider a
two stage offline-online framework,
where the offline
process is responsible for training a minimal set of AEs
from a collection of background images,
and the online process is aimed to evaluate 
the likelihood of changes (LoC) for each test image 
and finally determine whether each test image is change or no-change.
We assume
that
every query live image
is correctly paired with
a background image
beforehand
by using a global viewpoint information system such as GPS.
We do not assume
background images contain
only stationary objects,
but that dynamic objects may exist even in the background images.
We also do not assume
the query-background image pair
is perfectly registered,
but that a 
non-negligible number of registration errors may exist.

It is noteworthy that the objective of our method is different from and orthogonal to that of ensemble learning (EL) \cite{EL}. EL techniques, such as bagging and boosting, have been studied mainly in the context of alternative tasks of classification and clustering to reduce the dependence of the model on the specific dataset or data locality. As compared with that of these classical applications, the effective use of EL in AD is not straightforward because of (1) the unavailability of ground-truth training data and (2) the small sample space problem \cite{OutlierEnsembles}. The recently developed method presented in \cite{DBLP:journals/corr/SabokrouFFK16} is one of very few EL methods in which a set of AEs is employed for AD. However, the paper addressed a single AD problem, not multiple place-specific AD problems, as this paper does.

\subsection{Evaluating Likelihood of Change}

The basic idea of the online ICD process is
to reconstruct a query live image $I$
by using its counterpart (or linked) background model (i.e., AE) $c_{j}$, as shown in Fig. \ref{fig:C}.
The AE is designed to extract
the common factors of variation from normal samples
and reconstruct them accurately.
However, 
anomalous samples
do not contain these common factors of variation and thus
cannot be reconstructed accurately.
Therefore, 
the region-level LoC $V_{RE}$
for a given image region $P$ 
can be evaluated
by the RE at each pixel $p$:
\begin{equation}
V_{RE}(P) =
\sum_{p\in P} | I(p) - I'(p) |, \label{eq:re}
\end{equation}
where
the images $I$ and $I'$
are the input image and the image reconstructed by 
the AE
that is linked to the corresponding background image,
respectively,
and 
$|\cdot |$
is an absolute value operator.
If
$V_{RE}$
exceeds a pre-defined threshold $V_{RE}^*$ (Section \ref{sec:normalize}),
the interest region
$P$
is determined as an anomalous object.

\newcommand{\step}[2]{
\begin{flushleft}
    \fbox{
  \begin{minipage}[b]{16cm}
\begin{flushleft}
{\bf Step #1:} #2
\end{flushleft}
\end{minipage}
}
\end{flushleft}
}

\subsection{Training Autoencoders}

The rAE training procedure 
is described as follows.
We first initialize the training set $T_1$
with all the available training sets,
and set $i=1$.
Then, we iterate 
a process that consists of 
the following three steps.

\step{1}{
Given a training set $T_i$
we evaluate the RE (i.e., Eq. \ref{eq:re})
for each training image $I\in T_i$ 
using the current AE set $\{c_j\}_{j=1}^{N_i}$.
}

\step{2}{
The ID $j^*$ of the best AE that gives the minimum RE (i.e., Eq. \ref{eq:re}) is assigned to each training image $I$.
}

\step{3}{
If the RE value $V_{RE}$ is equal to or smaller than a pre-set threshold $V_{RE}^*$ for all the training images $T_i^+$,
the approximation accuracy is considered satisfactory
and the iteration is terminated.
Otherwise,
we 
return to the Step-1,
by setting
$i\leftarrow i+1$
and
$T\leftarrow T_{i}^+$,
where 
$T_{i}^+$
is a subset
of $T_i$
where the RE is greater than $V_{RE}^*$.
}

Our training procedure is partially
similar to that of OUTRES \cite{DBLP:conf/icde/MullerSS11} in its recursive nature.
Whereas most existing EL-based AD methods
construct a set of independent models from training data,
OUTRES is one of
a very few methods that construct models that 
are interdependent.
However,
our approach is different from 
OUTRES
in terms of its objective.
Whereas OUTRES uses the previous models for refining the current model,
we use them for virtual AD to
determine whether a new object is well explained
by the previous model.

\subsection{Accelerating Compression}

The proposed approach allows 
accelerated compression
of AEs.
Suppose we are given two independent
AE sets
$S$=$\{c_j\}_{j=1}^{N}$ 
and
our objective is to obtain a more compressed AE set
$S'$=$\{c_j'\}_{j=1}^{N'}$,
where
$N\ge N'$.
Our approach
allows us to use
the first
$N'$
AEs
in
the sequence $S$
(i.e., $\{c_j\}_{j=1}^{N'}$)
in place of the
$N'$
AEs
in $S'$.
As a result,
when we have a large AE set,
we can obtain
a smaller AE set,
without training a new AE.
This property
is important
in space limited systems,
such as
archiving and streaming applications.

\subsection{Normalizing Reconstruction Errors}\label{sec:normalize}

One design issue of the above training procedure 
is the determination of 
the
threshold $V_{RE}^*$
on $V_{RE}$
for different AEs.
Because
individual AEs are trained using
different training sets,
the RE outputs by different AEs
are not comparable.
To address this issue,
we propose normalizing
each RE value
by the AE-specific normalizer constant.

For the normalization,
we approximate
the probability distribution function (PDF)
of the REs
simply by a Gaussian distribution,
and normalize the RE value by
subtracting the mean value $\mu$
and dividing by 
its standard deviation (SD) $\sigma$
and by a normalizer coefficient $c$
set at $c=0.8$ in default.
This normalization
allows
outputs from different AEs
to be compared
and allows their direct comparison.

One desirable property of the SD-based normalizer
is that it can be updated incrementally
by incorporating new RE value $V_i$:
$\mu = N^{-1} \sum_{i=1}^N V_i$,
$\sigma^2 = (N^{-1} \sum_{i=1}^N V_i^2) - (N^{-1} \sum_{i=1}^N V_i)^2$.
This allows 
us
the
normal/anomalous
decision boundary
to be updated incrementally 
by incorporating new training images.

\subsection{Ranking Pixels}

We now consider the binary decision problem,
which takes
the list of pixels from all the test query images with their LoC values,
and
classifies each pixel as either change or no-change.
To achieve this,
we simply
sort the pixels
in descending order of the normalized LoC values:
the top-$X$ ranked pixels
are output as anomalous pixels.

\section{Experiments and Discussions}

We evaluated the proposed rAE-based ICD approach in terms of detection accuracy.

\subsection{Dataset}

We used the NCLT dataset \cite{nclt}. This dataset is a large-scale, long-term autonomy dataset for robotics research collected at the University of Michigan's North Campus by using a Segway vehicle platform (Fig. \ref{fig:nclt}). The data used in our study include view image sequences along the vehicle's trajectories acquired by the front facing camera of the Ladybug3.

\figD

From the viewpoint of the ICD benchmark, the NCLT dataset has desirable properties.
It includes
various types
of changing images
such as those of cars, pedestrians, building construction, construction machines, posters, and tables and whiteboards with wheels,
from seamless indoor and outdoor navigations of the Segway vehicle.
Moreover,
it
has recently
gained significant popularity
as a benchmark in
the robotics community  \cite{jmangelson-2018a}.

In the current study, we used four datasets ``2012/1/22," ``2012/3/31," ``2012/8/4," and ``2012/11/17" 
(hereafter, referred to as WI, SP, SU, and AU, respectively) collected across four different seasons, and annotated in the form of bounding boxes (BBs) 986 different changed objects in total that are found in all the possible 12 pairings of query and database seasons (i.e., $\{(i,j)|$$i,j\in\{WI,SP,SU,AU\}, i\neq j$$\}$). In addition, we prepared a collection of 1,973 random destructor images that do not contain changed objects and are independent of the 986 annotated images. Then, we merged these 1,973 destructor images and 986 annotated images to obtain a database of 2,959 images.

\subsection{Performance Metric}

Performance on
the ICD task
was evaluated in terms of
top-$X$ accuracy.
First,
we estimated the LoC image
by applying an ICD algorithm.
We then impose a 2D grid with $10\times 10$ pixel sized cells
on the query image
and estimated the LoC for each cell by max-pooling the pixel-wise LoC values from
all pixels 
that belong to that cell.
Next, all the cells from all query images were sorted
in descending order of LoC value,
and the
accuracies of the top-$X$ items in the list were evaluated.
We evaluated the top-$X$ accuracy
for different $X$ thresholds in
consideration of
the intersection-over-union (IoU) criterion \cite{yolo}.
For a specific $X$ threshold,
a successful detection was defined as a changed object, 
the annotated BB of which was sufficiently covered (IoU $\ge$ 50\% or 25\%) by the top-$X$ percent of cells.

\subsection{Comparing Methods}

We also compared the 
performance of the proposed method 
with that of two benchmark methods:
nearest neighbor anomaly detection (NN) \cite{accv12}, 
and 
AE with k-means -based place clustering (k-means) \cite{aeicd}.

The NN method
measures
the LoC of a query feature
by
its dissimilarity
to the most similar background feature.
First,
every query/background image
is represented by
a collection of SIFT features with Harris-Laplace keypoints \cite{sift}.
Then, the LoC at each keypoint in the query image
is measured by 
the L2 distance 
between a SIFT descriptor at that keypoint and its nearest-neighbor-SIFT in the database image.

The k-means method 
is different from the proposed method (AE) 
only in that 
it employs 
k-means clustering in the space of CNN features 
instead of the proposed clustering method.
Each of the $k$ trained AEs 
was used as the 
background model
of the background images that belong to the cluster.

\subsection{Performance Results}

Table \ref{tab:D} 
shows the performance results
for IoU$\ge 25$ and IoU$\ge 50$. 
We can see that 
the proposed method 
outperforms 
both the k-means and the NN
methods in
almost all the considered cases.
Among the compared methods,
the performance of the k-means's method was 
the second best.
NN
showed the poorest performance.
One reason for this is that
it produced
a number of small region proposals
and tended to be 
affected by minor appearance changes
between the query and background scenes.
In contrast,
the proposed method
succeeded in producing a 
precise pixel-wise anomaly score
by using the AE set was
trained to cover the 
collection of background images.

\tabD

\bibliographystyle{IEEEtran}
\bibliography{cite}

\end{document}